\def\year{2020}\relax
\documentclass[letterpaper]{article} 
\usepackage{aaai20}  
\usepackage{times}  
\usepackage{helvet} 
\usepackage{courier}  
\usepackage[hyphens]{url}  
\usepackage{graphicx} 
\usepackage{epsfig}
\usepackage{amsmath}
\usepackage{amsfonts,amssymb}
\usepackage{subcaption}
\usepackage{bm}
\usepackage{soul}
\usepackage{xcolor}
\usepackage[group-separator={,},group-minimum-digits={3}]{siunitx}
\usepackage{multicol}
\usepackage{booktabs}
\usepackage{algorithm}
\usepackage{algorithmicx}
\usepackage{algpseudocode}
\usepackage{colortbl}
\usepackage{dsfont}

\newcommand{\citet}[1]{\citeauthor{#1}~\shortcite{#1}}
\newcommand{\citep}{\cite}

\definecolor{shallow_red}{hsb}{0.15, 0.7, 0.95}
\definecolor{middle_red}{hsb}{0.11, 0.7, 0.95}
\definecolor{deep_red}{hsb}{0.07, 0.7, 0.95}

\title{Measuring and Relieving the Over-smoothing Problem for Graph Neural \\ Networks from the Topological View}
\author{Deli Chen,\textsuperscript{\rm 1}
Yankai Lin,\textsuperscript{\rm 2}
Wei Li,\textsuperscript{\rm 1}
Peng Li,\textsuperscript{\rm 2}
Jie Zhou,\textsuperscript{\rm 2}
Xu Sun\textsuperscript{\rm 1}
\\ 
\textsuperscript{\rm 1}{MOE Key Lab of Computational Linguistics, School of EECS, Peking University}\\
\textsuperscript{\rm 2}{Pattern Recognition Center, WeChat AI, Tencent Inc., China}\\
\{chendeli,liweitj47,xusun\}@pku.edu.cn,
\{yankailin,patrickpli,withtomzhou\}@tencent.com, 
}

\begin{document}
\maketitle
\begin{abstract}
Graph Neural Networks (GNNs) have achieved promising performance on a wide range of graph-based tasks. Despite their success, one severe limitation of GNNs is the over-smoothing issue (indistinguishable representations of nodes in different classes). In this work, we present a systematic and quantitative study on the over-smoothing issue of GNNs. 
First, we introduce two quantitative metrics, MAD and MADGap, to measure the smoothness and over-smoothness of the graph nodes representations, respectively.
Then, we verify that smoothing is the nature of GNNs and the critical factor leading to over-smoothness is the low information-to-noise ratio of the message received by the nodes, which is partially determined by the graph topology.
Finally, we propose two methods to alleviate the over-smoothing issue from the topological view: 
(1) MADReg which adds a MADGap-based regularizer to the training objective;
(2) AdaEdge which optimizes the graph topology based on the model predictions.
Extensive experiments on $7$ widely-used graph datasets with $10$ typical GNN models show that the two proposed methods are effective for relieving the over-smoothing issue, thus improving the performance of various GNN models. 
\end{abstract}

\section{Introduction}
\footnote{Accepted by AAAI 2020. This complete version contains the appendix.}Graph Neural Networks form an effective framework for learning graph representation, which have proven powerful in various graph-based tasks~\citep{dataset_ccp,model_gat,dataset_ppi}. Despite their success in graph modeling, over-smoothing is a common issue faced by GNNs~\citep{analysis_smoothing,survey_gnn}, which means that the representations of the graph nodes of different classes would become indistinguishable when stacking multiple layers, which seriously hurts the model performance (e.g., classification accuracy). However, there is limited study on explaining why and how over-smoothing happens.
In this work, we conduct a systematic and quantitative study of the over-smoothing issue of GNNs on $7$ widely-used graph datasets with $10$ typical GNN models, aiming to reveal what is the crucial factor bringing in the over-smoothing problem of GNNs and find out a reasonable direction to alleviate it. 

\begin{figure}[t!]
\centering
\includegraphics[scale=0.4]{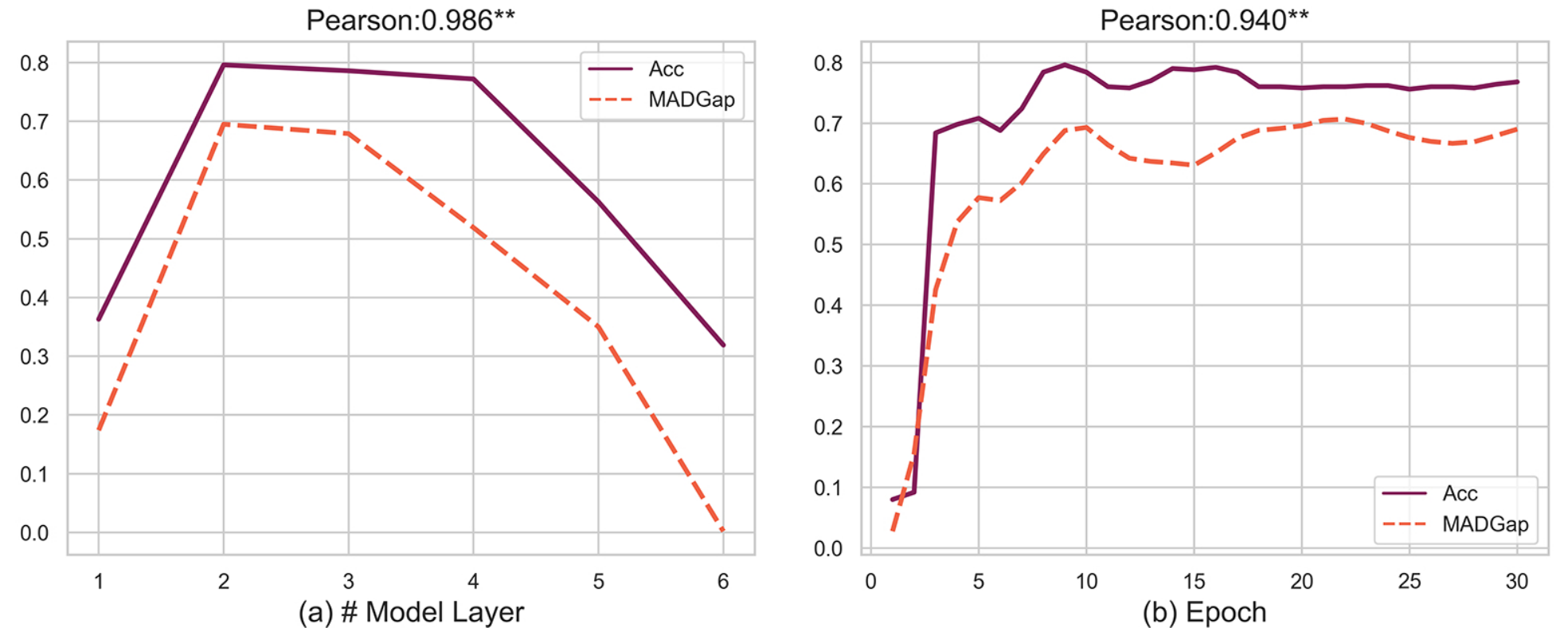}
\caption{The prediction accuracy (Acc) and MADGap of GCNs~\citep{model_gcn} on the \textit{CORA} dataset. We can observe a significantly high correlation between the accuracy and MADGap in two different situations: (a) different models: results of GCNs with different number of layers; (b) different training periods: results after each epoch in the 2-layer GCN. The Pearson correlation coefficient is shown in the title and ** means statistically significant with $p<0.01$.}
\label{figure_introduction}
\end{figure}

We first propose a quantitative metric Mean Average Distance (\textbf{MAD}), which calculates the mean average distance among node representations in the graph to measure the smoothness of the graph (\textbf{smoothness} means similarity of graph nodes representation in this paper). 
We observe that the MAD values of various GNNs become smaller as the number of GNN layers  increases, which supports the argument that smoothing is the essential nature of GNNs. Hence, the node interaction through the GNN message propagation would make their representations closer, and the whole graph representation would inevitably become smoothing when stacking multiple layers.

Furthermore, we argue that one key factor leading to the over-smoothing issue is the over-mixing of information and noise. 
The interaction message from other nodes may be either helpful information or harmful noise.
For example, in the node classification task, intra-class interaction can bring useful information, while inter-class interaction may lead to indistinguishable representations across classes. 
To measure the quality of the received message by the nodes, we define the \textbf{information-to-noise ratio} as the proportion of intra-class node pairs in all node pairs that have interactions through GNN model.
Based on our hypothesis, we extend MAD to MADGap to measure the over-smoothness of graph (\textbf{over-smoothness} means similarity of representations among different classes' nodes in this paper). We notice that two nodes with close topological distance (can reach with a few hops) are more likely to belong to the same class, and vice versa. 
Therefore, we differentiate the role between remote and neighboring nodes and calculate the gap of MAD values (\textbf{MADGap}) between remote and neighboring nodes to estimate the over-smoothness of graph representation.
Experimental results prove that MADGap does have a significantly high correlation with the model performance in general situations, and an example is shown in Figure~\ref{figure_introduction}. 
Further experiments show that both the model performance and the MADGap value rise as the information-to-noise ratio increases, which verifies our assumption that the information-to-noise ratio affects the smoothness of graph representation to a great extent.

After more in-depth analysis, we propose that low information-to-noise ratio is caused by the discrepancy between the graph topology and the objective of the downstream task. In the node classification task, if there are too many inter-class edges, the nodes will receive too much message from nodes of other classes after several propagation steps, which would result in over-smoothing. To prove our assumption, we optimize the graph topology by removing inter-class edges and adding intra-class edges based on the gold labels, which proves very effective in relieving over-smoothing and improving model performance. 
Hence, the graph topology has a great influence on the smoothness of graph representation and model performance. That is to say, there is a deviation from the natural graph to the downstream task. However, in the previous graph-related studies~\citep{model_gat,model_ggnn,model_arma}, researchers mainly focus on designing novel GNN architectures but pay less attention to improve the established graph topology.

Based on our observations, we propose two methods to relieve the over-smoothing issue from the topological view: (a) \textbf{MADReg:} we add a MADGap-based regularizer to the training objective to directly increase received information and reduce noise; (b) Adaptive Edge Optimization (\textbf{AdaEdge}\footnote{The algorithm name is changed from the previous AdaGraph to AdaEdge since the conflicting using of AdaGraph with other work.}): we iteratively train GNN models and conduct edge remove/add operations based on the prediction to adjust the graph adaptively for the learning target. Experimental results show that our two proposed methods can significantly relieve the over-smoothing issue and improve model performance in general cases, which further verifies our conclusions and provides a compelling perspective towards better GNNs performance.

The contributions of this work are threefold:
\begin{itemize}
    \item We conduct a systematic and quantitative study of the over-smoothing issue on a wide range of graph datasets and models. We propose and verify that a key factor behind the over-smoothing issue is the information-to-noise ratio which is influenced by the graph topology.
    \item We design two quantitative metrics: MAD for smoothness and MADGap for over-smoothness of graph representation.  Statistical analysis shows that MADGap has a significantly high correlation with model performance. 
    \item We propose two methods: MADReg and AdaEdge to relieve the over-smoothing issue of GNNs. Experimental results show that our proposed methods can significantly reduce over-smoothness and improve the performance of multiple GNNs on various datasets.
\end{itemize}

\section{Datasets and Models}
node classification task, one of the most basic graph-based tasks, is usually conducted to verify the effectiveness of  GNN architectures~\citep{model_gat,model_sage} or analyze the characteristics of GNNs~\citep{analysis_smoothing,analysis_lowpass}.
Therefore, we select the node classification task for our experiments. We conduct experiments on $7$ public datasets in three types, namely, (1) citation network: \textit{CORA}, \textit{CiteSeer}, \textit{PubMed}~\citep{dataset_real_ccp}; (2) coauthor network: CS, Physics;\footnote{\url{https://kddcup2016.azurewebsites.net}} (3) Amazon product network: Computers, Photo~\citep{dataset_real_amazon}. 
We conduct our detailed analysis on the three citation networks, which are usually taken as the benchmarks for graph-related studies~\citep{analysis_smoothing,analysis_lowpass} and verify the effectiveness of the proposed method on all these datasets. 

To guarantee the generalizability of our conclusion, we conduct experiments with $10$ typical GNN models in this work. 
The GNN models and their propagation methods are listed in Table~\ref{table_model}, in which the propagation taxonomy follows~\citet{survey_gnn}.
The implementation of the baselines is partly based on~\citet{code_geometric} and~\citet{analysis_lowpass}. 
More details about the datasets and experimental settings are given in Appendix~A.

\begin{table}[t]
\centering
\resizebox{.9\columnwidth}{!}{
\begin{tabular}{ll}
\toprule
\textbf{Model}  & \textbf{Propagate}                                                         \\
\midrule
\textbf{GCN}~\citep{model_gcn} & Convolution                                                       \\
\textbf{ChebGCN}~\citep{model_cheb}        & Convolution                \\                              \textbf{HyperGraph}~\citep{model_hyper_graph}    
       & \begin{tabular}[c]{@{}l@{}}Convolution\\ \&Attention\end{tabular} \\
\textbf{FeaSt}~\citep{model_feast}       & Convolution          \\
\textbf{GraphSAGE}~\citep{model_sage}       & Convolution        \\
\textbf{GAT}~\citep{model_gat}     & Attention            \\
\textbf{ARMA}~\citep{model_arma}       & Convolution          \\
\textbf{GraphSAGE}~\citep{model_sage}       & Convolution         \\
\textbf{HighOrder}~\citep{model_stand_graph}       & Attention            \\
\textbf{GGNN}~\citep{model_ggnn}       & Gated              \\        
\bottomrule
\end{tabular}}
\caption{Introduction of baseline GNN models. The information propagation method is also displayed.}
\label{table_model}
\end{table}

\section{Measuring Over-smoothing Problem from the Topological View}
In this section, we aim to investigate what is the key factor leading to the over-smoothing problem. To this end, we propose two quantitative metrics MAD and MADGap to measure the smoothness and over-smoothness of graph representation, which are further used to analyze why and how the over-smoothing issue happens.
\subsection{MAD: Metric for Smoothness}
To measure the smoothness of the graph representation, we first propose a quantitative metric: Mean Average Distance (MAD). MAD reflects the smoothness of graph representation by calculating the mean of the average distance from nodes to other nodes. Formally, given the graph representation matrix $\bm{H} \in \mathbb{R}^{n \times h}$ (we use the hidden representation of the final layer of GNN. Term $h$ is the hidden size), we first obtain the distance matrix $\bm{D} \in \mathbb{R}^{n\times n}$ for $\bm{H}$ by computing the cosine distance between each node pair:
\begin{equation}
    D_{ij} = 1 - \frac{\bm{H}_{i,:} \cdot \bm{H}_{j,:}}{|\bm{H}_{i,:}| \cdot |\bm{H}_{j,:}|}\qquad i,j \in [1,2,\cdots,n],
\end{equation}
where $\bm{H}_{k,:}$ is the $k$-th row of $\bm{H}$.
The reason to use cosine distance is that cosine distance is not affected by the absolute value of the node vector, thus better reflecting the smoothness of graph representation. 
Then we filter the target node pairs by element-wise multiplication $\bm{D}$ with a mask matrix $\bm{M}^{tgt}$
\begin{equation}
 \bm{D}^{tgt} = \bm{D} \circ \bm{M}^{tgt},
\end{equation}
where $\circ$ denotes element-wise multiplication; $\bm{M}^{tgt}\in \{0,1\}^{n\times n}$; $\bm{M}^{tgt}_{ij}=1$ only if node pair $(i,j)$ is the target one. Next we access the average distance $\bar{\bm{D}}^{tgt}$ for non-zero values along each row in $\bm{D}^{tgt}$:
\begin{equation}
    \bar{\bm{D}}^{tgt}_i = \frac{\sum_{j=0}^n \bm{D}^{tgt}_{ij}}{\sum_{j=0}^n \mathds{1}\left( {\bm{D}}^{tgt}_{ij}\right)},
\end{equation}
where $\mathds{1}(x)=1$ if $x>0$ otherwise $0$.
Finally, the MAD value given the target node pairs is calculated by averaging the non-zero values in $\bar{\bm{D}}^{tgt}$:
\begin{eqnarray}
 \mathrm{MAD^{tgt}} &=& \frac{\sum_{i=0}^n \bar{\bm{D}}^{tgt}_{i}}{\sum_{i=0}^n \mathds{1}\left(\bar {\bm{D}}^{tgt}_{i}\right)}.
\end{eqnarray}

\citet{analysis_smoothing} perform a theoretical analysis on the graph convolution network (GCN), and conclude that performing smoothing operation on node representations is the key mechanism why GCN works.
We extend the conclusion empirically to the $10$ typical GNNs listed in Table~\ref{table_model} with the help of the proposed MAD metric.
To this end, for each GNN model with different number of layers, we compute the MAD value $\mathrm{MAD^{global}}$ by taking all node pairs into account, i.e., all values in $\bm{M}^{tgt}$ are $1$, to measure the global smoothness of the learned graph representation.

\begin{figure}[t]
\centering
\includegraphics[scale=0.6]{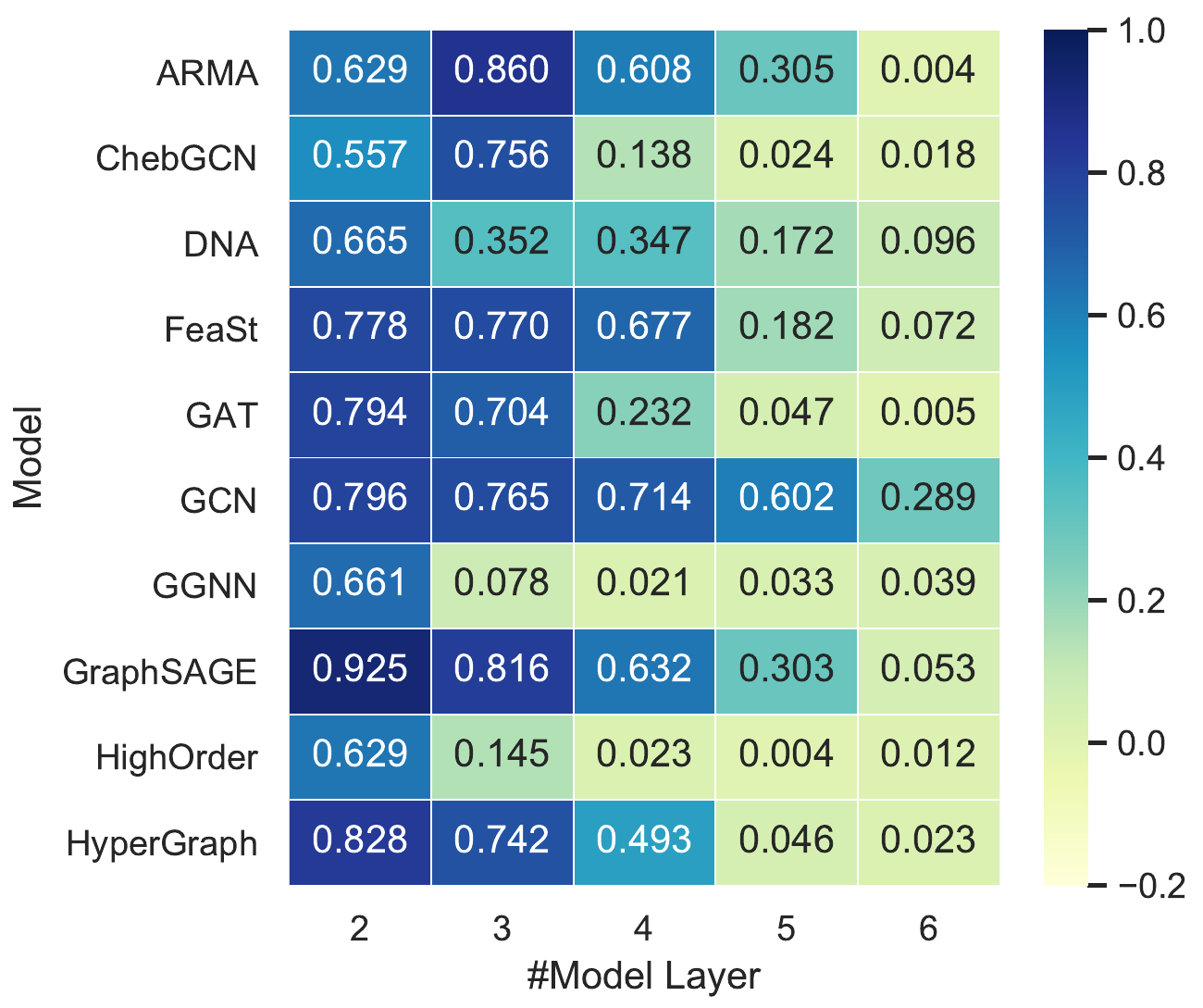}
\caption{The MAD values of various GNNs with different layers on the \textit{CORA} dataset. Darker color means larger MAD value. We can find that the smoothness of graph representation rises as the model layer increases.}
\label{figure_smooth_pattern}
\end{figure}

The results on the \textit{CORA} dataset are shown in Figure~\ref{figure_smooth_pattern}. We can observe that as the number of GNN layers increases, the MAD values become smaller. 
Apart from this, the MAD value of high-layer GNNs gets close to $0$, which means that all the node representations become indistinguishable. 
GNN models update the node representation based on the features from neighboring nodes. We observe that the interaction between nodes makes their representations similar to each other. 
Similar phenomenons that the smoothness rises as the layer increases are also observed in other datasets as presented in Appendix~B. Therefore, we conclude that smoothing is an essential nature for GNNs. 

\begin{figure}
\centering
\includegraphics[width=0.8\columnwidth]{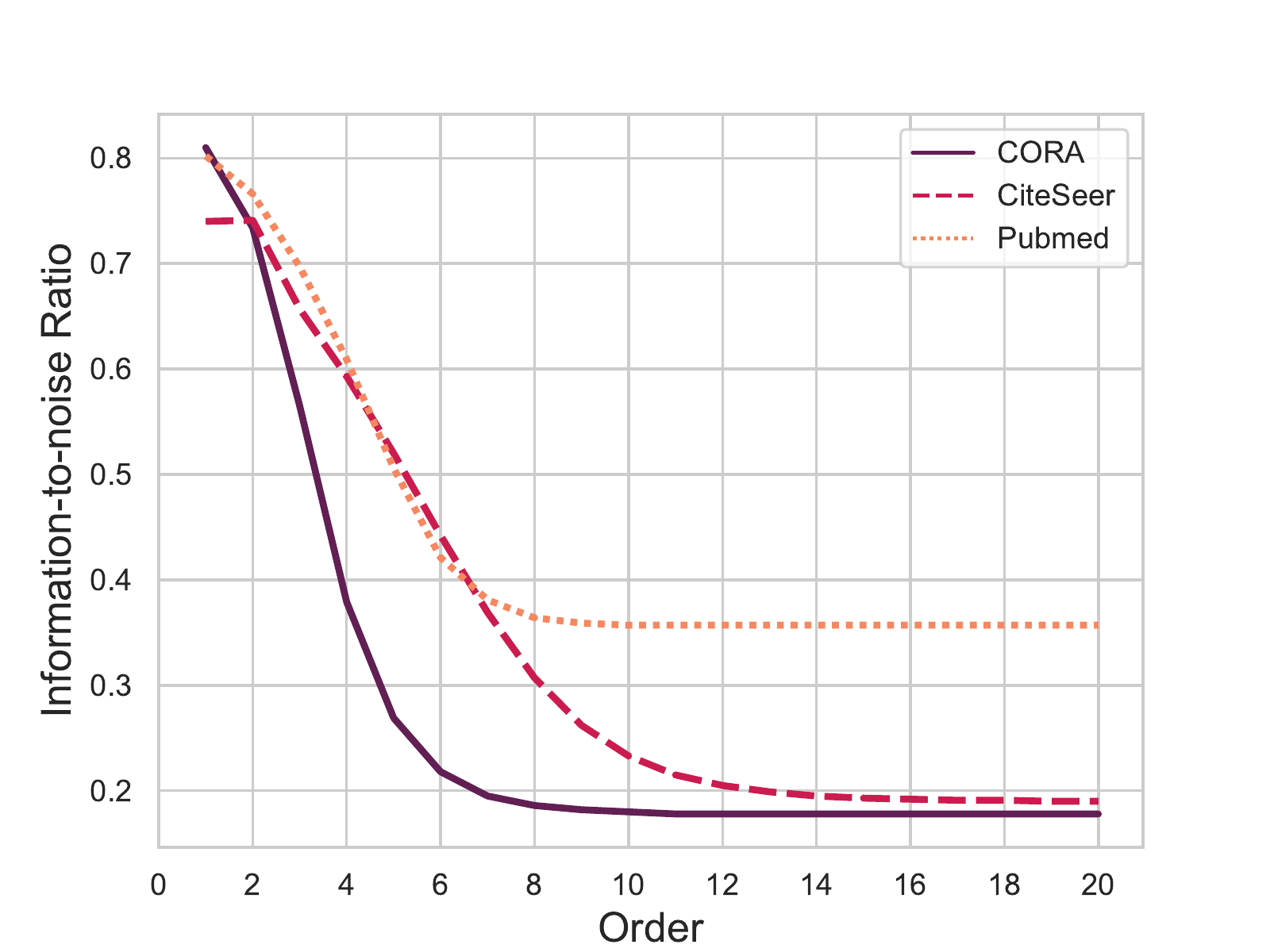}
\caption{The information-to-noise ratio at different neighbor orders (accumulated) for the \textit{CORA/CiteSeer/PubMed} datasets. We can find that the information-to-noise ratio declines as the orders increases in all these three datasets.}
\label{figure_type_order}
\end{figure}

\begin{figure*}
\centering
\includegraphics[width=0.98\textwidth]{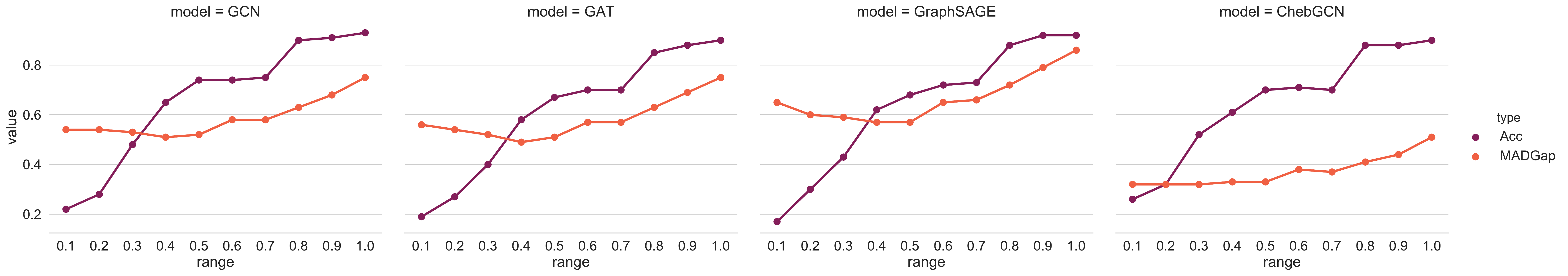}
\caption{Performance (accuracy) and over-smoothness (MADGap) of node sets with different information-to-noise ratio (e.g., $0.1$ means ratio$\le0.1$) on the \textit{CORA} dataset (We display 4 out of 10 models results due to the limited space. We observe similar results in other models). All models have 2 layers. Results prove that nodes with higher information-to-noise ratio would have less over-smoothness degree and better prediction result.}
\label{figure_info_noise}
\end{figure*}

\subsection{Information-to-noise Ratio Largely Affects Over-smoothness}

With the help of MAD, we can quantitatively measure the smoothness of graph representation. Here come two new questions: since smoothing is the nature of GNNs, what is over-smoothing, and what results in over-smoothing?

We assume that the over-smoothing problem is caused by the over-mixing of information and noise, which is influenced by the quality of the nodes received message.
The interaction message from other nodes by GNN operation may be either helpful information or interference noise.
For example, in the node classification task, interaction between nodes of the same class brings useful information, which makes their representations more similar to each other and the probability of being classified into the same class is increased. 
On the contrary, the contact of nodes from other classes brings the noise. Hence, the reason why GNNs work is that the received useful information is more than noise. On the other hand, when the noise is more than the information, the learned graph representation will become over-smoothing.

To quantitatively measure the quality of the received message of the nodes, we define the information-to-noise ratio as the proportion of intra-class node pairs in all contactable node pairs that have interactions through the GNN model. For example, at the second-order, the information-to-noise ratio for each node is the proportion of nodes of the same class in all the first-order and second-order neighbors; the information-to-noise ratio for the whole graph is the proportion of the intra-class pairs in all the node pairs that can be contacted in 2 steps.
In Figure~\ref{figure_type_order}, we display the information-to-noise ratio of the whole graph for the \textit{CORA, CiteSeer} and \textit{Pubmed} datasets.
We can find that there are more intra-class node pairs at low order and vice versa. 
When the model layer number gets large where the information-to-noise ratio is small, the interaction between high-order neighbors brings too much noise and dilutes the useful information, which is the reason for the over-smoothing issue.
Based on this observation, we extend MAD to MADGap to measure the over-smoothness in the graph representation. 
From Figure~\ref{figure_type_order} we notice that two nodes with small topological distance (low-order neighbours) are more likely to belong to the same category. Hence, we propose to utilize the graph topology to approximate the node category, and calculate the gap of MAD values differentiating remote and neighbour nodes to estimate the over-smoothness of the graph representation,
\begin{equation}
\mathrm{MADGap} =  \mathrm{MAD^{rmt}} -  \mathrm{MAD^{neb}},
\end{equation}
where $\mathrm{MAD^{rmt}}$ is the $\mathrm{MAD}$ value of the remote nodes in the graph topology and $\mathrm{MAD^{neb}}$ is the $\mathrm{MAD}$ value of the neighbouring nodes.

According to our assumption, large MADGap value indicates that the useful information received by the node is more than noise. At this time, GNNs perform reasonable extent of smoothing, and the model would perform well. On the contrary, small or negative MADGap means over-smoothing and inferior performance. 
To verify the effectiveness of MADGap, we calculate the MADGap value\footnote{In this work, we calculate $\mathrm{MAD^{neb}}$ based on nodes with orders $\le3$ and $\mathrm{MAD^{rmt}}$ based on nodes with orders $\ge8$.} and compute the Pearson coefficient between the MADGap and the prediction accuracy for various GNN models. 
We report the Pearson coefficient for GNNs with different layers on \textit{CORA}, \textit{CiteSeer} and \textit{PubMed} datasets in Table~\ref{table_madgap_pearson}.
According to the table, we can find that there exists a significantly high correlation between MADGap and the model performance, which validates that MADGap is a reliable metric to measure graph representation over-smoothness. Besides, MADGap can also be used as an observation indicator to estimate the model performance based on the graph topology without seeing the gold label.
It is worth noting that 1-layer GNN usually has small MADGap and prediction accuracy (Figure~\ref{figure_introduction}), which is caused by the insufficient information transfer, while the over-smoothing issue of high-layer GNN is caused by receiving too much noise.
 
In Figure~\ref{figure_info_noise}, we show the MADGap and prediction accuracy for node sets with different information-to-noise ratios in the same model. We can find that even with the same model and propagation step, nodes with higher rate of information-to-noise ratio generally have higher prediction accuracy with smaller over-smoothing degree. We also observe similar phenomena on other datasets, which are shown in Appendix~C. This way, we further verify that it is the information-to-noise ratio that affects the graph representation over-smoothness to a great extent, thus influencing the model performance.

\begin{table}[t]
	\centering
	\small
	\resizebox{.85\columnwidth}{!}{
		\begin{tabular}{l|lll}
			\hline
			\textbf{Model }      & \textbf{CORA}    & \textbf{CiteSeer} & \textbf{PubMed }  \\ \hline
			\textbf{GCN }        & 0.986$^{**}$ & 0.948$^{**}$  & 0.971$^{**}$ \\
			\textbf{ChebGCN }    & 0.945$^{**}$ & 0.984$^{**}$  & 0.969$^{**}$ \\
			\textbf{HyperGraph } & 0.990$^{**}$ & 0.965$^{**}$  & 0.932$^{**}$ \\
			\textbf{FeaSt }      & 0.993$^{**}$ & 0.986$^{**}$  & 0.906$^{*}$ \\
			\textbf{GraphSAGE }  & 0.965$^{**}$ & 0.995$^{*}$  & 0.883   \\
			\textbf{GAT }        & 0.960$^{**}$ & 0.998$^{**}$  & 0.965$^{**}$ \\
			\textbf{ARMA }       & 0.909$^{*}$ & 0.780    & 0.787   \\
			\textbf{HighOrder }  & 0.986$^{**}$ & 0.800    & 0.999$^{**}$ \\
			\textbf{DNA }        & 0.945$^{**}$ & 0.884$^{*}$  & 0.887$^{*}$ \\
			\textbf{GGNN }       & 0.940$^{*}$ & 0.900$^{*}$  & 0.998$^{**}$ \\ \hline
	\end{tabular}}
	\caption{The Pearson coefficient between accuracy and MADGap for various models on \textit{CORA/CiteSeer/PubMed} datasets. Pearson coefficient is calculated based on the results of models with different layers (1-6). * means statistically significant with $p<0.05$ and ** means $p<0.01$.}
	\label{table_madgap_pearson}
\end{table}

\begin{figure*}[t]
\centering
\includegraphics[width=0.97\textwidth]{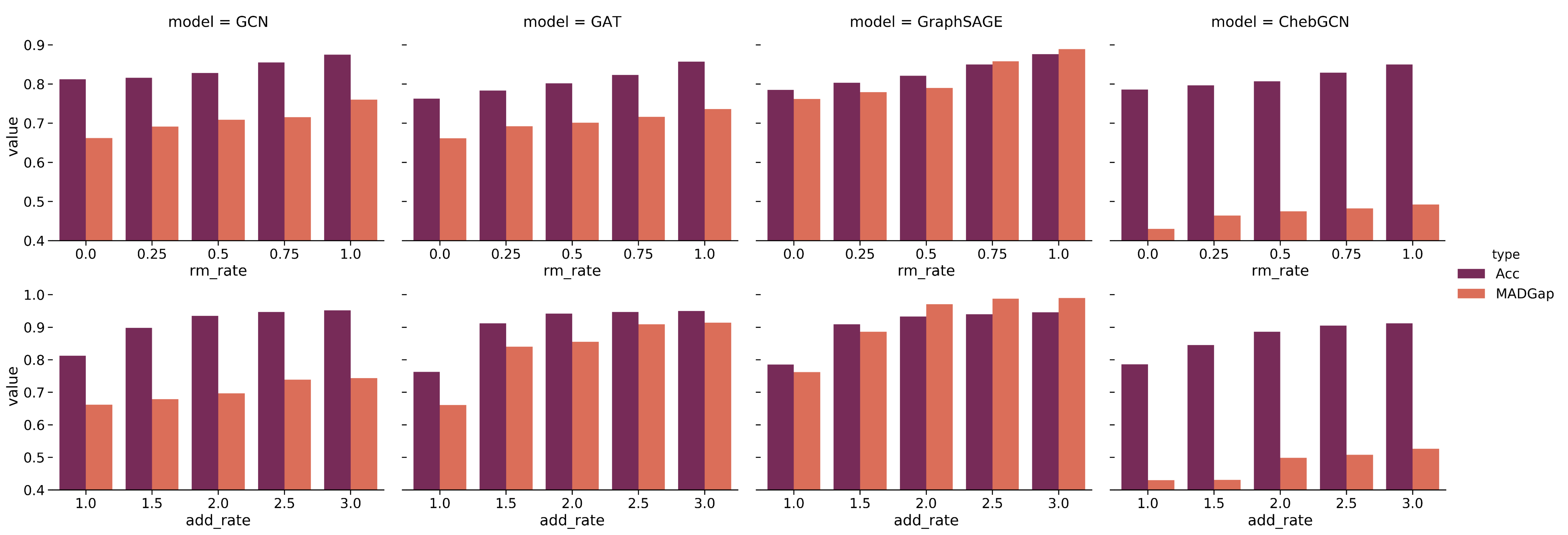}
\caption{The gold label based topology adjustment experiment on the \textit{CORA} dataset. We show the results of both removing inter-class edges (first row, where the X-axis represents the removing rate) and adding intra-class edges (second row, where the X-axis represents the intra-class edge ratio compared to the raw graph) on GCN, GAT, GraphSAGE and ChebGCN. Results show that both of these methods are very helpful for relieving the over-smoothing issue and improving model performance.}
\label{figure_golden_adjust}
\end{figure*}

\subsection{Topology Affects the Information-to-noise Ratio}
From the previous analysis, we can find that the key factor influencing the smoothness of graph representation is the information-to-noise ratio.
Then the following question is: what affects the information-to-noise ratio?
We argue that it is the graph topology that affects the information-to-noise ratio. The reason for the node receiving too much noise is related to the discordance between the natural graph and the task objective.
Take node classification as an example. If there are too many inter-class edges, the nodes will receive too much noise after multiple steps of message propagation, which results in over-smoothing and bad performance. 

The graph topology is constructed based on the natural links. For example, the edges in the citation network represent the citing behavior between papers and edges in the product network represent the products co-purchasing relations.  
GNN models rely on these natural links to learn node representations. However, natural links between nodes of different classes are harmful to the node classification task. 
Therefore, we propose to alleviate the over-smoothing issue of GNNs and improve their performance by optimizing the graph topology to match the downstream task.

To verify our assumption, we optimize the graph topology by removing inter-class edges and adding intra-class edges based on the gold labels. The results on the  \textit{CORA} dataset are shown in Figure~\ref{figure_golden_adjust}.  We can find that the MADGap value rises consistently as more inter-class edges are removed and more intra-class edges are added, resulting in better model performance. Therefore, optimizing graph topology is helpful in relieving the over-smoothing problem and improving model performance.

In summary, we find that the graph topology has a great influence on the smoothness of graph representation and model performance. 
However, there is still discordance between the natural links and the downstream tasks. 
Most existing works mainly focus on designing novel GNN architectures but pay less attention to the established graph topology. 
Hence, we further investigate to improve the performance of GNNs by optimizing the graph topology.

\begin{figure*}[t]
\centering
\includegraphics[width=0.99\textwidth]{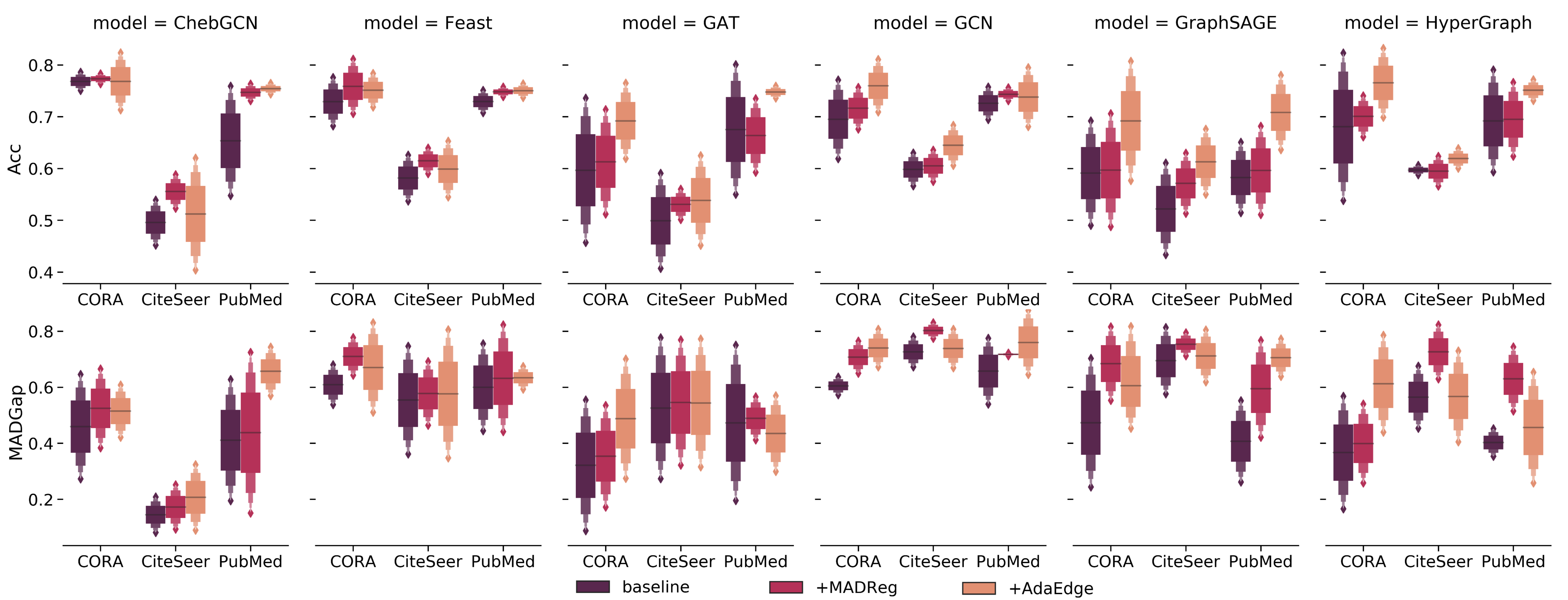}
\caption{MADReg and AdaEdge results on the \textit{CORA/CiteSeer/PubMed} datasets. The number of GNN layers is 4, where the over-smoothing issue is severe. The box plot shows the mean value and the standard deviation of the prediction accuracy and the MADGap values of 50 turns results (5 dataset splitting methods and 10 random seeds for each splitting following~\citet{dataset_amazon} and~\citet{method_fishergcn}. More details can be found in Appendix~A). And we can find that the two proposed methods can effectively relieve the over-smoothing issue and improve model performance in most cases.}
\label{figure_box}
\end{figure*}

\section{Relieving Over-smoothing Problem from the Topological View}
Inspired by the previous analysis, we propose two methods to relieve the over-smoothing issue from the topological view: 
(1) \textbf{MADReg:} we add a MADGap-based regularizer to the training objective; 
(2) Adaptive Edge Optimization (\textbf{AdaEdge}): we adjust the graph topology adaptively by iteratively training GNN models and conducting edge remove/add operations based on the prediction result.
Neither of these two methods is restricted to specific model architectures and can be used in the training process of general GNN models. Experiments demonstrate their effectiveness in a variety of GNNs. 

\subsection{MADReg: MADGap as Regularizer}
In the previous experiments, we find that MADGap shows a significantly high correlation with model performance. 
Hence, we add MADGap to the training objective to make the graph nodes receive more useful information and less interference noise:
\begin{equation}
 \mathcal{L} = \sum -l\,\log\,p(\,\hat{l}\,|\bm{X},\bm{A},\Theta) - \lambda\mathrm{MADGap},
\end{equation}
where $\bm{X}$ is the input feature matrix, $\bm{A}$ is the adjacency matrix, $\hat{l}$ and $l$ are the predicted and gold labels of the node respectively. $\Theta$ is the parameters of GNN and $\lambda$ is the regularization coefficient to control the influence of MADReg. 
We calculate $\mathrm{MADGap}$ on the training set to be consistent with the cross-entropy loss.

\subsection{AdaEdge: Adaptive Edge Optimization}
As discussed in the previous section, after optimizing the topology based on gold label (adding the intra-class edges and removing the inter-class edges), the over-smoothing issue is notably alleviated, and the model performance is greatly improved. Inspired by this, we propose a self-training algorithm called AdaEdge to optimize the graph topology based on the prediction result of the model to adaptively adjust the topology of the graph to make it more reasonable for the specific task objective. Specifically, we first train GNN on the original graph and adjust the graph topology based on the prediction result of the model by deleting inter-class edges and adding intra-class edges. Then we re-train the GNN model on the updated graph from scratch. We perform the above graph topology optimization operation multiple times. 
The details of the AdaEdge algorithm are introduced in Appendix~D. 

\subsection{Relieving Over-smoothing in High-order Layers}

To verify the effectiveness of the two proposed methods, we conduct controlled experiments for all the $10$ baseline GNN models on \textit{CORA/CiteSeer/PubMed} datasets. We calculate the prediction accuracy and MADGap value for the GNN models with $4$ layers, where the over-smoothing issue is serious. The results are shown in Figure~ \ref{figure_box}. We present 6 out of 10 models results due to the space limit; the other models can be found in Appendix~E.
We can find that in the high-order layer situation where the over-smoothing issue is severe, the MADReg and AdaEdge methods can effectively relieve the over-smoothing issue and improve model performance for most models in all three datasets.
The effectiveness of MADReg and AdaEdge further validates our assumption and provides a general and effective solution to relieve the over-smoothing problem.

\begin{table*}[t]
\centering
\small
\resizebox{2.1\columnwidth}{!}{
\begin{tabular}{l|ll|ll|ll|ll|ll|ll|ll}
\hline
 \textbf{Acc(\%)}         & {\bf CORA} & & {\bf CiteSeer} & & {\bf PubMed} &  & \multicolumn{2}{l|}{ \textbf{ Amazon Photo}} & \multicolumn{2}{l|}{ \textbf{ Amazon Comp.}} & \multicolumn{2}{l|}{ \textbf{ Coauthor CS}} & \multicolumn{2}{l}{ \textbf{ Coauthor Phy.}}   \\ \hline
Model      & baseline        & +AE            & baseline        & +AE         & baseline           & +AE      & baseline        & +AE             & baseline           & +AE             & baseline             & +AE            & baseline              & +AE              \\ \hline
GCN        
& 81.2\tiny$\pm$0.8        & \cellcolor{deep_red}{$82.3^{**}_{\pm0.8}$}       
& 69.3\tiny$\pm$0.7        & \cellcolor{shallow_red}{$69.7^{**}_{\pm0.9}$}
& 76.3\tiny$\pm$0.5        & \cellcolor{deep_red}{$77.4^{**}_{\pm0.5}$}        
& 90.6\tiny$\pm$0.7        & \cellcolor{middle_red}{$91.5^{**}_{\pm0.5}$}        
& 81.7\tiny$\pm$0.7        & \cellcolor{middle_red}{$82.4^{**}_{\pm1.1}$}     
& 89.8\tiny$\pm$0.3        & \cellcolor{middle_red}{$90.3^{**}_{\pm0.4}$}       
& 92.8\tiny$\pm$1.6         & \cellcolor{shallow_red}{$93.0^{**}_{\pm1.1}$}
         \\ 
ChebGCN       
& 78.6\tiny$\pm$0.6        & {$80.1^{**}_{\pm0.5}$}        
& 67.4\tiny$\pm$1.0        & \cellcolor{shallow_red}{$67.8^{*}_{\pm1.2}$}    
& 76.7\tiny$\pm$0.1        & \cellcolor{middle_red}{$77.5^{**}_{\pm0.6}$}      
& 89.6\tiny$\pm$1.6        & {$89.4^{*}_{\pm1.2}$}        
& 80.8\tiny$\pm$2.4        & \cellcolor{middle_red}{$81.3^{**}_{\pm1.1}$}        
& 90.5\tiny$\pm$0.4        & \cellcolor{shallow_red}{$90.7^{*}_{\pm0.3}$}      
& \textbackslash{} & \textbackslash{}
           \\ 
HyperGraph 
& 80.5\tiny$\pm$0.6        & \cellcolor{middle_red}{$81.4^{**}_{\pm0.8}$}     
& 67.9\tiny$\pm$0.5        & \cellcolor{middle_red}{$68.5^{**}_{\pm0.5}$} 
& 77.4\tiny$\pm$0.2        & {$77.3_{\pm0.7}$}      
& 87.5\tiny$\pm$0.7        & \cellcolor{deep_red}{$88.6^{**}_{\pm0.3}$}        
& 58.7\tiny$\pm$22.1       & \cellcolor{deep_red}{$61.4^{**}_{\pm26.8}$}       
& 86.9\tiny$\pm$0.5        &\cellcolor{shallow_red}{$87.3^{**}_{\pm0.4}$}        
& 91.9\tiny$\pm$2.0         & \cellcolor{shallow_red}{$92.2^{**}_{\pm1.4}$}
       \\ 
FeaSt      
& 80.4\tiny$\pm$0.7        & \cellcolor{deep_red}{$81.6^{**}_{\pm0.7}$}        
& 69.3\tiny$\pm$1.1        & \cellcolor{shallow_red}{$69.4^{*}_{\pm1.0}$}   
& 76.6\tiny$\pm$0.6        & \cellcolor{middle_red}{$77.2^{*}_{\pm0.4}$} 
& 90.5\tiny$\pm$0.6        & \cellcolor{shallow_red}{$90.8^{**}_{\pm0.6}$}        
& 80.8\tiny$\pm$1.3        & \cellcolor{middle_red}{$81.7^{**}_{\pm0.9}$}        
& 88.4\tiny$\pm$0.2        & \cellcolor{middle_red}{$88.9^{**}_{\pm0.2}$}      
& \textbackslash{}   & \textbackslash{}
           \\ 
GraphSAGE  
& 78.5\tiny$\pm$1.7        & \cellcolor{middle_red}{$80.2^{**}_{\pm1.2}$}      
& 68.4\tiny$\pm$0.9        & \cellcolor{deep_red}{$69.4^{**}_{\pm0.8}$}
& 75.2\tiny$\pm$1.1        & \cellcolor{deep_red}{$77.2^{**}_{\pm0.8}$}      
& 90.1\tiny$\pm$1.4        & \cellcolor{middle_red}{$90.6^{**}_{\pm0.5}$}       
& 80.2\tiny$\pm$1.0        & \cellcolor{middle_red}{$81.1^{**}_{\pm1.0}$}      
& 90.1\tiny$\pm$0.4        & \cellcolor{shallow_red}{$90.3^{**}_{\pm0.4}$}    
& 93.0\tiny$\pm$0.4        & {$92.7_{\pm0.2}$} 
    \\ 
GAT        
& 76.3\tiny$\pm$3.1        & \cellcolor{deep_red}{$77.9^{**}_{\pm2.0}$}       
& 68.9\tiny$\pm$0.6        & \cellcolor{shallow_red}{$69.1^{*}_{\pm0.8}$}
& 75.9\tiny$\pm$0.5        & \cellcolor{deep_red}{$76.6^{**}_{\pm0.2}$}       
 & 89.7\tiny$\pm$1.7        & \cellcolor{deep_red}{$90.8^{**}_{\pm0.9}$}        
& 81.4\tiny$\pm$1.5        & {$81.1^{*}_{\pm1.6}$}        
& 85.5\tiny$\pm$1.9        & \cellcolor{deep_red}{$86.6^{**}_{\pm1.6}$}       
& 91.1\tiny$\pm$1.0         &\cellcolor{shallow_red}{$91.4^{*}_{\pm1.0}$} 
     \\ 
ARMA       
& 74.9\tiny$\pm$10.6       & \cellcolor{deep_red}{$76.4^{**}_{\pm5.6}$}        
& 65.3\tiny$\pm$4.1        & \cellcolor{middle_red}{$66.1^{**}_{\pm4.3}$} 
& 68.5\tiny$\pm$11.4        & \cellcolor{middle_red}{$68.9^{*}_{\pm12.2}$}        
& 86.4\tiny$\pm$3.0        & \cellcolor{middle_red}{$87.0^{**}_{\pm1.9}$}        
& 63.8\tiny$\pm$18.9       & \cellcolor{deep_red}{$71.7^{**}_{\pm8.1}$}       
& 90.6\tiny$\pm$1.1        & \cellcolor{shallow_red}{$90.9^{**}_{\pm0.6}$}      
& 92.2\tiny$\pm$1.8         & \cellcolor{shallow_red}{$92.6^{*}_{\pm1.0}$}
          \\ 
HighOrder  
& 76.6\tiny$\pm$1.2        & {$72.5^{**}_{\pm4.1}$}        
& 64.2\tiny$\pm$1.0        & {$63.3^{**}_{\pm1.0}$}
& 75.0\tiny$\pm$2.6        & \cellcolor{deep_red}{$76.9^{**}_{\pm1.3}$}         
& 26.1\tiny$\pm$12.4       & \cellcolor{deep_red}{$30.3^{**}_{\pm10.2}$}       
& 26.3\tiny$\pm$12.7       & {$23.9^{*}_{\pm13.4}$}       
& 84.2\tiny$\pm$1.0        &\cellcolor{deep_red}{$85.6^{**}_{\pm0.7}$}       
& 90.8\tiny$\pm$0.8         & \cellcolor{shallow_red}{$90.9_{\pm0.6}$}
          \\ 
DNA        
& 58.2\tiny$\pm$14.4       & \cellcolor{deep_red}{$60.1^{*}_{\pm10.8}$}          
& 60.9\tiny$\pm$2.7        & \cellcolor{shallow_red}{$61.3^{**}_{\pm2.2}$} 
& 65.8\tiny$\pm$7.8        & \cellcolor{deep_red}{$66.8^{*}_{\pm9.6}$} 
& 89.1\tiny$\pm$1.3        & \cellcolor{middle_red}{$89.8^{**}_{\pm0.6}$}       
& 78.2\tiny$\pm$2.9        & \cellcolor{deep_red}{$79.8^{**}_{\pm2.0}$}        
& 88.2\tiny$\pm$0.9        & \cellcolor{deep_red}{$90.0^{**}_{\pm0.6}$}       
& 93.0\tiny$\pm$0.5         & \cellcolor{shallow_red}{$93.3^{*}_{\pm0.4}$}       
         \\ 
GGNN       
& 47.3\tiny$\pm$6.1        & {$44.7^{**}_{\pm3.5}$}        
& 55.5\tiny$\pm$2.8        & {$47.9^{**}_{\pm3.4}$} 
& 66.1\tiny$\pm$4.4        & \cellcolor{deep_red}{$69.5^{**}_{\pm1.2}$}   
& 74.1\tiny$\pm$12.3       & \cellcolor{deep_red}{$80.6^{**}_{\pm7.2}$}        
& 42.4\tiny$\pm$26.7       & \cellcolor{deep_red}{$61.5^{**}_{\pm20.8}$}      
& 86.6\tiny$\pm$1.4        & \cellcolor{deep_red}{$88.2^{**}_{\pm0.8}$}     
& 91.2\tiny$\pm$1.2         & \cellcolor{shallow_red}{$91.6^{**}_{\pm0.7}$}      
     \\ \hline
\end{tabular}}
\caption{Controlled experiments of AdaEdge (+AE) on all the 7 datasets. We show the mean value, the standard deviation and the t-test significance of 50 turns results. 
* means statistically significance with $p<0.05$ and ** means $p<0.01$. Darker color means larger improvement. The missing results are due to the huge consumption of GPU memory of large graphs.}
\label{table_loop_result1}
\end{table*}

\subsection{Improving Performance of GNNs}
In Table~\ref{table_loop_result1}, we show the controlled experiments for GNN models trained on the original graph and the updated graph obtained by the AdaEdge method on all the $7$ datasets. We select the best hyper-parameters when training GNN on the original graph and fix all these hyper-parameters when training on the updated graph. 
Experimental results show that the AdaEdge method can effectively improve the model performance in most cases, which proves that optimizing the graph topology is quite helpful for improving model performance. 
We analyze the cases of the AdaEdge method with little or no improvement and find that this is caused by the incorrect operations when adjusting the topology. 
Therefore, when the ratio of incorrect operations is too large, it will bring serious interference to the model training and bring in little or no improvement.
Due to the space limit, the results of MADReg are shown in Appendix~F. 
Typically, the baselines achieve their best performance with small number of GNN layers, where the over-smoothing issue is not severe. Under this condition, MADReg can hardly improve the performance by enlarging the MADGap value. However, when the over-smoothing issue becomes more severe while the GNN layer number grows larger, MADReg is still capable of improving the performance of the baselines significantly.
Above all, both AdaEdge and MADReg are effective for improving GNNs performance, and AdaEdge generalizes better when the over-smoothing issue is not severe.

\section{Related Work}
\subsection{Graph Neural Networks (GNNs)}
GNNs have proven effective in various non-Euclidean graph structures, such as social network~\citep{model_sage}, biology network~\citep{dataset_ppi}, business graph~\citep{dataset_real_amazon} and academic graph~\citep{dataset_real_ccp}. 
Recently, many novel GNN architectures have been developed for graph-based tasks.~\citet{model_gat} propose the graph attention network to use self-attention to aggregate information from neighboring nodes.~\citet{model_sage} propose a general inductive framework to generate node embedding by sampling and aggregating features from the neighboring nodes. There are also other GNNs proposed, such as ARMA~\citep{model_arma}, FeaSt~\citep{model_feast}, HyperGraph~\citep{model_hyper_graph} and so on.~\citet{jump_knowledge} propose jumping knowledge networks to help the GNN model to leverage the information from high-order neighbours for a better node representation.
However, all these models focus on improving the information propagation and aggregation operation on the static graph while paying less attention to the graph topology. 
In this work, we propose to explicitly optimize the graph topology to make it more suitable for the downstream task.

\citet{dynamic_evolvegcn} propose the EvolveGCN that uses the RNN to evolve the graph model itself over time.~\citet{model_dna} allow for a selective and node-adaptive aggregation of the  neighboring embeddings of potentially differing locality.~\citet{model_togcn} propose a new variation of GCN by jointly refining the topology and training the fully connected network.
These existing works about dynamic graph rely on the adaptive ability of the model itself and focus on special GNN architecture (e.g., GCN),  while our AdaEdge method optimizes the graph topology with a clear target (adding intra-class edges and removing inter-class edges) and can be used in general GNN architectures.~\citet{dropedge} propose DropEdge method to drop edges randomly at each training epoch for data augmentation while our AdaEdge method adjusts edges before training to optimize the graph topology.

\subsection{Over-smoothing Problem in GNNs}
Previous works~\citep{analysis_smoothing,survey_gnn} have proven that over-smoothing is a common phenomenon in GNNs.
\citet{analysis_smoothing} prove that the graph convolution of the GCN model is actually a special form of Laplacian smoothing.
~\citet{smooth_useful} propose that smoothness is helpful for node classification and design methods to encourage the smoothness of the output distribution, while~\citet{smooth_noise} propose that nodes may be mis-classified by topology based attribute smoothing and try to overcome this issue.
In this work, we prove that smoothing is the essential feature of GNNs, and then classify the smoothing into two kinds by the information-to-noise ratio: reasonable smoothing that makes GNN work, and over-smoothing that causes the bad performance.
From this view, the methods from~\citet{smooth_useful} and~\citet{smooth_noise} can be regarded as improving reasonable smoothing and relieve over-smoothing, respectively.
Besides,~\citet{liwei2019recursive} propose to use LSTM in GNN to solve over-smoothing issue in text classification. However, existing works usually mention the over-smoothing phenomenon, but there lacks systematic or quantitative research about it. 

\section{Conclusion and Future Work}
In this work, we conduct a systematic and quantitative study of the over-smoothing issue faced by GNNs. We first design two quantitative metrics: MAD for smoothness and MADGap for over-smoothness. 
From the quantitative measurement results on multiple GNNs and graph datasets, we find that smoothing is the essential nature of GNNs; over-smoothness is caused by the over-mixing of information and the noise.
Furthermore, we find that there is a significantly high correlation between the MADGap and the model performance. Besides, we prove that the information-to-noise ratio is related to the graph topology, and we can relieve the over-smoothing issue by optimizing the graph topology to make it more suitable for downstream tasks. 
Followingly, we propose two methods to relieve the over-smoothing issue in GNNs: the MADReg and the AdaEdge methods. Extensive results prove that these two methods can effectively relieve the over-smoothing problem and improve model performance in general situations.

Although we have shown optimizing graph topology is an effective way of improving GNNs performance, our proposed AdaEdge method still suffers from the wrong graph adjustment operation problem. How to reduce these operations is a promising research direction. 

\section{Acknowledgement}
This work was supported in part by a Tencent Research Grant and National Natural Science Foundation of China (No. 61673028). Xu Sun is the corresponding author of this paper.

\bibliography{aaai2020}

\begin{thebibliography}{}

\bibitem[\protect\citeauthoryear{Bai, Zhang, and
  Torr}{2019}]{model_hyper_graph}
Bai, S.; Zhang, F.; and Torr, P.~H.
\newblock 2019.
\newblock Hypergraph {Convolution and Hypergraph Attention}.
\newblock {\em arXiv preprint arXiv:1901.08150}.

\bibitem[\protect\citeauthoryear{Bianchi \bgroup et al\mbox.\egroup
  }{2019}]{model_arma}
Bianchi, F.~M.; Grattarola, D.; Livi, L.; and Alippi, C.
\newblock 2019.
\newblock {Graph Neural Networks with Convolutional Arma Filters}.
\newblock {\em arXiv preprint arXiv:1901.01343}.

\bibitem[\protect\citeauthoryear{Defferrard, Bresson, and
  Vandergheynst}{2016}]{model_cheb}
Defferrard, M.; Bresson, X.; and Vandergheynst, P.
\newblock 2016.
\newblock {Convolutional Neural Networks on Graphs with Fast Localized Spectral
  Filtering}.
\newblock In {\em Advances in Neural Information Processing Systems},
  3837--3845.

\bibitem[\protect\citeauthoryear{Deng, Dong, and Zhu}{2019}]{smooth_useful}
Deng, Z.; Dong, Y.; and Zhu, J.
\newblock 2019.
\newblock Batch virtual adversarial training for graph convolutional networks.
\newblock {\em arXiv preprint arXiv:1902.09192}.

\bibitem[\protect\citeauthoryear{Fey and Lenssen}{2019}]{code_geometric}
Fey, M., and Lenssen, J.~E.
\newblock 2019.
\newblock {Fast Graph Representation Learning with {PyTorch Geometric}}.
\newblock In {\em ICLR Workshop on Representation Learning on Graphs and
  Manifolds}.

\bibitem[\protect\citeauthoryear{Fey}{2019}]{model_dna}
Fey, M.
\newblock 2019.
\newblock {Just Jump: Dynamic Neighborhood Aggregation in Graph Neural
  Networks}.
\newblock {\em arXiv preprint arXiv:1904.04849}.

\bibitem[\protect\citeauthoryear{Hamilton, Ying, and
  Leskovec}{2017}]{model_sage}
Hamilton, W.~L.; Ying, Z.; and Leskovec, J.
\newblock 2017.
\newblock {Inductive Representation Learning on Large Graphs}.
\newblock In {\em Advances in Neural Information Processing Systems},
  1024--1034.

\bibitem[\protect\citeauthoryear{Kipf and Welling}{2017}]{model_gcn}
Kipf, T.~N., and Welling, M.
\newblock 2017.
\newblock {Semi-supervised Classification with Graph Convolutional Networks}.
\newblock In {\em 5th International Conference on Learning Representations,
  {ICLR} 2017}.

\bibitem[\protect\citeauthoryear{Li \bgroup et al\mbox.\egroup
  }{2016}]{model_ggnn}
Li, Y.; Tarlow, D.; Brockschmidt, M.; and Zemel, R.~S.
\newblock 2016.
\newblock {Gated Graph Sequence Neural Networks}.
\newblock In {\em 4th International Conference on Learning Representations,
  {ICLR} 2016}.

\bibitem[\protect\citeauthoryear{Li \bgroup et al\mbox.\egroup
  }{2019a}]{liwei2019recursive}
Li, W.; Li, S.; Ma, S.; He, Y.; Chen, D.; and Sun, X.
\newblock 2019a.
\newblock Recursive graphical neural networks for text classification.
\newblock {\em arXiv preprint arXiv:1909.08166}.

\bibitem[\protect\citeauthoryear{Li \bgroup et al\mbox.\egroup
  }{2019b}]{graph2seq_liweiacl}
Li, W.; Xu, J.; He, Y.; Yan, S.; Wu, Y.; and Sun, X.
\newblock 2019b.
\newblock {Coherent Comment Generation for Chinese Articles with a
  Graph-to-Sequence Model}.
\newblock In {\em Proceedings of the 57th Conference of the Association for
  Computational Linguistics},  4843--4852.

\bibitem[\protect\citeauthoryear{Li, Han, and Wu}{2018}]{analysis_smoothing}
Li, Q.; Han, Z.; and Wu, X.-M.
\newblock 2018.
\newblock {Deeper Insights into Graph Convolutional Networks for
  Semi-supervised Learning}.
\newblock In {\em Thirty-Second AAAI Conference on Artificial Intelligence}.

\bibitem[\protect\citeauthoryear{Maehara}{2019}]{analysis_lowpass}
Maehara, T.
\newblock 2019.
\newblock Revisiting graph neural networks: All we have is low-pass filters.
\newblock {\em arXiv preprint arXiv:1905.09550}.

\bibitem[\protect\citeauthoryear{McAuley \bgroup et al\mbox.\egroup
  }{2015}]{dataset_real_amazon}
McAuley, J.; Targett, C.; Shi, Q.; and Van Den~Hengel, A.
\newblock 2015.
\newblock {Image-based Recommendations on Styles and Substitutes}.
\newblock In {\em Proceedings of the 38th International ACM SIGIR Conference on
  Research and Development in Information Retrieval},  43--52.
\newblock ACM.

\bibitem[\protect\citeauthoryear{Morris \bgroup et al\mbox.\egroup
  }{2019}]{model_stand_graph}
Morris, C.; Ritzert, M.; Fey, M.; Hamilton, W.~L.; Lenssen, J.~E.; Rattan, G.;
  and Grohe, M.
\newblock 2019.
\newblock {Weisfeiler and Leman go Neural: Higher-order Graph Neural Networks}.
\newblock In {\em Proceedings of the AAAI Conference on Artificial
  Intelligence}, volume~33,  4602--4609.

\bibitem[\protect\citeauthoryear{Pareja \bgroup et al\mbox.\egroup
  }{2019}]{dynamic_evolvegcn}
Pareja, A.; Domeniconi, G.; Chen, J.; Ma, T.; Suzumura, T.; Kanezashi, H.;
  Kaler, T.; and Leisersen, C.~E.
\newblock 2019.
\newblock Evolvegcn: Evolving graph convolutional networks for dynamic graphs.
\newblock {\em arXiv preprint arXiv:1902.10191}.

\bibitem[\protect\citeauthoryear{Rong \bgroup et al\mbox.\egroup
  }{2019}]{dropedge}
Rong, Y.; Huang, W.; Xu, T.; and Huang, J.
\newblock 2019.
\newblock The truly deep graph convolutional networks for node classification.
\newblock {\em arXiv preprint arXiv:1907.10903}.

\bibitem[\protect\citeauthoryear{Sen \bgroup et al\mbox.\egroup
  }{2008}]{dataset_real_ccp}
Sen, P.; Namata, G.; Bilgic, M.; Getoor, L.; Galligher, B.; and Eliassi-Rad, T.
\newblock 2008.
\newblock Collective classification in network data.
\newblock {\em AI magazine} 29(3):93--93.

\bibitem[\protect\citeauthoryear{Shchur \bgroup et al\mbox.\egroup
  }{2018}]{dataset_amazon}
Shchur, O.; Mumme, M.; Bojchevski, A.; and G{\"u}nnemann, S.
\newblock 2018.
\newblock {Pitfalls of Graph Neural Network Evaluation}.
\newblock {\em arXiv preprint arXiv:1811.05868}.

\bibitem[\protect\citeauthoryear{Sun, Koniusz, and
  Wang}{2019}]{method_fishergcn}
Sun, K.; Koniusz, P.; and Wang, J.
\newblock 2019.
\newblock {Fisher-Bures Adversary Graph Convolutional Networks}.
\newblock In {\em Proceedings of the Thirty-Fifth Conference on Uncertainty in
  Artificial Intelligence},  161.

\bibitem[\protect\citeauthoryear{Veli{\v{c}}kovi{\'c} \bgroup et
  al\mbox.\egroup }{2018}]{model_gat}
Veli{\v{c}}kovi{\'c}, P.; Cucurull, G.; Casanova, A.; Romero, A.; Lio, P.; and
  Bengio, Y.
\newblock 2018.
\newblock {Graph Attention Networks}.
\newblock In {\em 6th International Conference on Learning Representations,
  {ICLR} 2018}.

\bibitem[\protect\citeauthoryear{Verma, Boyer, and Verbeek}{2018}]{model_feast}
Verma, N.; Boyer, E.; and Verbeek, J.
\newblock 2018.
\newblock {Feastnet: Feature-steered Graph Convolutions for 3d Shape Analysis}.
\newblock In {\em Proceedings of the IEEE Conference on Computer Vision and
  Pattern Recognition},  2598--2606.

\bibitem[\protect\citeauthoryear{Xu \bgroup et al\mbox.\egroup
  }{2018}]{jump_knowledge}
Xu, K.; Li, C.; Tian, Y.; Sonobe, T.; Kawarabayashi, K.; and Jegelka, S.
\newblock 2018.
\newblock Representation learning on graphs with jumping knowledge networks.
\newblock In {\em Proceedings of the 35th International Conference on Machine
  Learning, {ICML} 2018},  5449--5458.

\bibitem[\protect\citeauthoryear{Yang \bgroup et al\mbox.\egroup
  }{2019a}]{smooth_noise}
Yang, L.; Chen, Z.; Gu, J.; and Guo, Y.
\newblock 2019a.
\newblock {Dual Self-Paced Graph Convolutional Network: Towards Reducing
  Attribute Distortions Induced by Topology}.
\newblock In {\em Proceedings of the Twenty-Eighth International Joint
  Conference on Artificial Intelligence, {IJCAI} 2019},  4062--4069.

\bibitem[\protect\citeauthoryear{Yang \bgroup et al\mbox.\egroup
  }{2019b}]{model_togcn}
Yang, L.; Kang, Z.; Cao, X.; Jin, D.; Yang, B.; and Guo, Y.
\newblock 2019b.
\newblock {Topology Optimization based Graph Convolutional Network}.
\newblock In {\em Proceedings of the Twenty-Eighth International Joint
  Conference on Artificial Intelligence, {IJCAI} 2019},  4054--4061.

\bibitem[\protect\citeauthoryear{Yang, Cohen, and
  Salakhutdinov}{2016}]{dataset_ccp}
Yang, Z.; Cohen, W.~W.; and Salakhutdinov, R.
\newblock 2016.
\newblock {Revisiting Semi-supervised Learning with Graph Embeddings}.
\newblock In {\em Proceedings of the 33nd International Conference on Machine
  Learning,{ICML} 2016},  40--48.

\bibitem[\protect\citeauthoryear{Zhou \bgroup et al\mbox.\egroup
  }{2018}]{survey_gnn}
Zhou, J.; Cui, G.; Zhang, Z.; Yang, C.; Liu, Z.; and Sun, M.
\newblock 2018.
\newblock {Graph Neural Networks: A Review of Methods and Applications}.
\newblock {\em arXiv preprint arXiv:1812.08434}.

\bibitem[\protect\citeauthoryear{Zitnik and Leskovec}{2017}]{dataset_ppi}
Zitnik, M., and Leskovec, J.
\newblock 2017.
\newblock {Predicting Multicellular Function through Multi-layer Tissue
  Networks}.
\newblock {\em Bioinformatics} 33(14):i190--i198.

\end{thebibliography}
\bibliographystyle{aaai}

\appendix
\section{A Experimental Settings}
In this section, we will introduce the graph task datasets and the baseline GNN models considered in this work first. And then we will introduce the experiment details and notations used in this paper.

\begin{table}[ht]
\centering
\resizebox{.85\columnwidth}{!}{
\begin{tabular}{l|rrrr}
\hline
 & \textbf{Class} & \textbf{Feature} & \textbf{Node} & \textbf{Edge}  \\ \hline
\textbf{CORA}             & 7                & 1,433              & 2,708           & 5,278                 \\
\textbf{CiteSeer}         & 6                & 3,703              & 4,732           & 13,703                   \\
\textbf{PubMed}           & 3                & 500                 & 44,338          & 500                      \\
\textbf{Coauthor CS}      & 15               & 6,805               & 18,333          & 81,894                    \\
\textbf{Coauthor Phy.} & 5                & 8,415                & 34,493          & 247,962                    \\
\textbf{Amazon Comp.} & 10               & 767              & 13,381          & 245,778                    \\
\textbf{Amazon Photo}     & 8                & 745                & 7,487           & 119,043                     \\\hline
\end{tabular}}
\caption{Statistical information about datasets.}
\label{table_dataset}
\end{table}

\begin{table*}[t]
\centering
\small
\resizebox{1.65\columnwidth}{!}{
\begin{tabular}{l|l|ll|ll|ll}
\hline
      &         & \textbf{CORA }        &             & \textbf{CiteSeer }    &             & \textbf{PubMed}    &           \\ \hline
  LAYER    &   Metric      & GCN         & +MADReg     & GCN         & +MADReg     & GCN       & +MADReg   \\ \hline
Layer1 & acc(\%) 
      & 54.6\tiny \tiny$\pm$14    & \cellcolor{shallow_red}{$54.9^{*}_{\pm13}$}     
      & 39.7\tiny$\pm$12          & \cellcolor{deep_red}{$41.6^{**}_{\pm12}$}  
      & 52.8\tiny$\pm$20          & \cellcolor{deep_red}{$57.1^{**}_{\pm16}$}   \\
      & MADGap  
      & 0.13\tiny$\pm$0.04   & {$0.14^{*}_{\pm0.04}$}
      & 0.07\tiny$\pm$0.04   & {$0.10^{**}_{\pm0.03}$}
      & 0.33\tiny$\pm$0.20   & {$0.44^{**}_{\pm0.21}$}           \\ \hline
Layer2 & acc(\%) 
      & 80.8\tiny$\pm$0.8    & {$80.6_{\pm0.88}$}
      & 68.7\tiny$\pm$0.96   & {$67.1^{**}_{\pm1.7}$}   
      & 76.6\tiny$\pm$0.66   & {$75.2^{**}_{\pm1.2}$}  \\
      & MADGap  
      & 0.59\tiny$\pm$0.03   & {$0.66^{**}_{\pm0.04}$} 
      & 0.51\tiny$\pm$0.08   & {$0.53^{*}_{\pm0.07}$}
      & 0.74\tiny$\pm$0.03   & {$0.73_{\pm0.05}$}        \\ \hline
Layer3 & acc(\%) 
      & 79.0\tiny$\pm$1.2    & \cellcolor{middle_red}{$79.9^{**}_{\pm1.1}$}   
      & 65.9\tiny$\pm$1.6    & {$65.4^{*}_{\pm2.2}$}   
      & 76.2\tiny$\pm$0.65   & \cellcolor{shallow_red}{$76.5^{*}_{\pm0.64}$} \\
      & MADGap  
      & 0.57\tiny$\pm$0.03   & {$0.59^{*}_{\pm0.04}$}
      & 0.52\tiny$\pm$0.04   & {$0.53^{*}_{\pm0.03}$} 
      & 0.70\tiny$\pm$0.04   & {$0.74^{**}_{\pm0.07}$}        \\ \hline
Layer4 & acc(\%) 
      & 70.8\tiny$\pm$12     & \cellcolor{deep_red}{$72.1^{**}_{\pm8.2}$}    
      & 55.6\tiny$\pm$11     & \cellcolor{deep_red}{$62.2^{**}_{\pm1.9}$}   
      & 67.3\tiny$\pm$11     & \cellcolor{deep_red}{$69.4^{**}_{\pm9.9}$}  \\
      & MADGap  
      & 0.52\tiny$\pm$0.05   & {$0.53^{*}_{\pm0.04}$}
      & 0.47\tiny$\pm$0.07   & {$0.54^{**}_{\pm0.04}$} 
      & 0.63\tiny$\pm$0.22   & {$0.71^{**}_{\pm0.13}$}       \\ \hline
Layer5 & acc(\%) 
      & 46.9\tiny$\pm$9.7    & \cellcolor{deep_red}{$48.0^{**}_{\pm11}$}   
      & 35.8\tiny$\pm$10     & \cellcolor{deep_red}{$46.5^{**}_{\pm7.7}$}    
      & 55.1\tiny$\pm$9.4    & \cellcolor{deep_red}{$56.1^{**}_{\pm9.6}$}  \\
      & MADGap  
      & 0.24\tiny$\pm$0.07   & {$0.28^{*}_{\pm0.04}$}
      & 0.331\tiny$\pm$0.18  & {$0.46^{**}_{\pm0.07}$} 
      & 0.45\tiny$\pm$0.22   & {$0.51^{**}_{\pm0.24}$}       \\ \hline
Layer6 & acc(\%) 
      & 36.2\tiny$\pm$9.1    & \cellcolor{deep_red}{$38.5^{**}_{\pm8.2}$}
      & 28.9\tiny$\pm$9.0    & \cellcolor{deep_red}{$37.5^{**}_{\pm4.7}$}   
      & 49.5\tiny$\pm$9.1    & \cellcolor{deep_red}{$52.3^{**}_{\pm10.5}$}   \\
      & MADGap  
      & -0.22\tiny$\pm$0.20  & {$-0.20^{**}_{\pm0.16}$}
      & 0.18\tiny$\pm$0.23   & {$0.36^{**}_{\pm0.07}$}
      &   0.12\tiny$\pm$0.25 &{$0.18^{*}_{\pm0.24}$}           \\ \hline
\end{tabular}}
\caption{Results of MADReg method for GCN with different layers on \textit{CORA/CiteSeer/PubMed} dataset. Darker color means larger improvement over the baseline.}

\label{table_reg_result2}
\end{table*}

\subsection{Experiment Dataset}
We conduct experiment on $7$ node classification datasets in $3$ types:
\begin{itemize}
    \item \textbf{Citation Network} The \textit{CORA}, \textit{CiteSeer}, \textit{PubMed} datasets~\citep{dataset_real_ccp} are citation networks which are usually taken as be benchmarks of graph related studies~\citep{analysis_smoothing,analysis_lowpass}. The nodes are papers in computer science field with bag-of-word features of paper title. The edges represent the citation relation amaong papers and the label is paper category.
    \item \textbf{Coauthor Network} Coauthor CS and Coauthor Physics are co-authorship graphs based on the Microsoft Academic Graph from the KDD Cup 2016 challenge 3.\footnote{\url{https://kddcup2016.azurewebsites.net} } Here, nodes are authors, that are connected by an edge if they co-authored a paper; node features represent paper keywords for each author’s papers, and class labels indicate most active fields of study for each author.
    \item \textbf{Amazon Network} Amazon Computers and Amazon Photo~\citep{dataset_real_amazon} are segments of the Amazon co-purchase graph, where nodes represent goods, edges indicate that two goods are frequently bought together, node features are bag-of-words encoded product reviews, and class labels are given by the product category.
\end{itemize}
Datasets related statistical information are shown  in the Table~\ref{table_dataset}. All these graphs are undirected graphs with no edge weight.

\subsection{Experimental Details}
Previous works~\citep{model_gat,model_hyper_graph,graph2seq_liweiacl} on GNN study usually run experiment multi-turns to eliminate random effects. And \citet{dataset_amazon} pointed out that in the semi-supervised node classification task, the train/valid/test split of dataset has a significant influence on the final result. 
Following~\citet{dataset_amazon,method_fishergcn} we apply the 20/30/rest splitting method, which means we randomly sample 20 nodes in each category for training set and 30 for validation set; all the rest nodes are as test data.
In order to ensure the credibility of the results, we select 5 random train/valid/test split of dataset and run 10 turns with different random seeds in each split. Then we measure the average number of all the 50 turns' results.

Besides, in order to avoid the random effects caused by dataset split or initial seeds and observe the influence of the proposed methods more clearly, we use the same random dataset split and initial seed list for the baseline method and the proposed method in each controlled experiment, and the dataset split and seed list was randomly generated before each controlled experiment. We also fix all the other hyper-parameters in each controlled experiment. 

\subsection{Notations Used in This Paper}
Given a undirected graph $\bm{G}  = \left ( \bm{V},\bm{E} \right )$, where the $\bm{V}$ is the node set with $|\bm{V}|=n $ and $\bm{E}$ is the edge set. 
The adjacency matrix is denoted by $\bm{A} = [a_{ij}] \in \mathbb{R}^{n\times n}$. The raw node features are denoted by $\bm{X} = [\bm{x_1},\bm{x_2},\dots,\bm{x_n}]^\top \in \mathbb{R}^{n \times c}$ and each $\bm{x_i} \in \mathbb{R}^{c}$ is the input feature vector for the $i$-th node. 
Each node has a corresponding label $l $ indicating the node class. In the semi-supervised node classification task, the gold labels of nodes in the trainset are known and the target is to predict the labels for nodes in the testset.
We use $\mathbf{GNN_k}$ to represent a k-layer (propagate step) graph neural network and the label predicted by this GNN is represented by $ \hat{l}$. Besides, the hidden states of all node after $j$ layer(s) is denoted by $\bm{H_j} \in \mathbb{R}^{n\times h_j}$ ($h_j$ represents the hidden-size of GNN $j$-th layer) and $\bm{h_{ji}}$ denotes the hidden state vector of the $i$-th node representation after $j$ layers.

\section{B MAD Global Values on More Datasets }
\begin{figure}[t]
\centering
\small
\begin{minipage}{.48\textwidth}
  \centering
  \includegraphics[width=\linewidth]{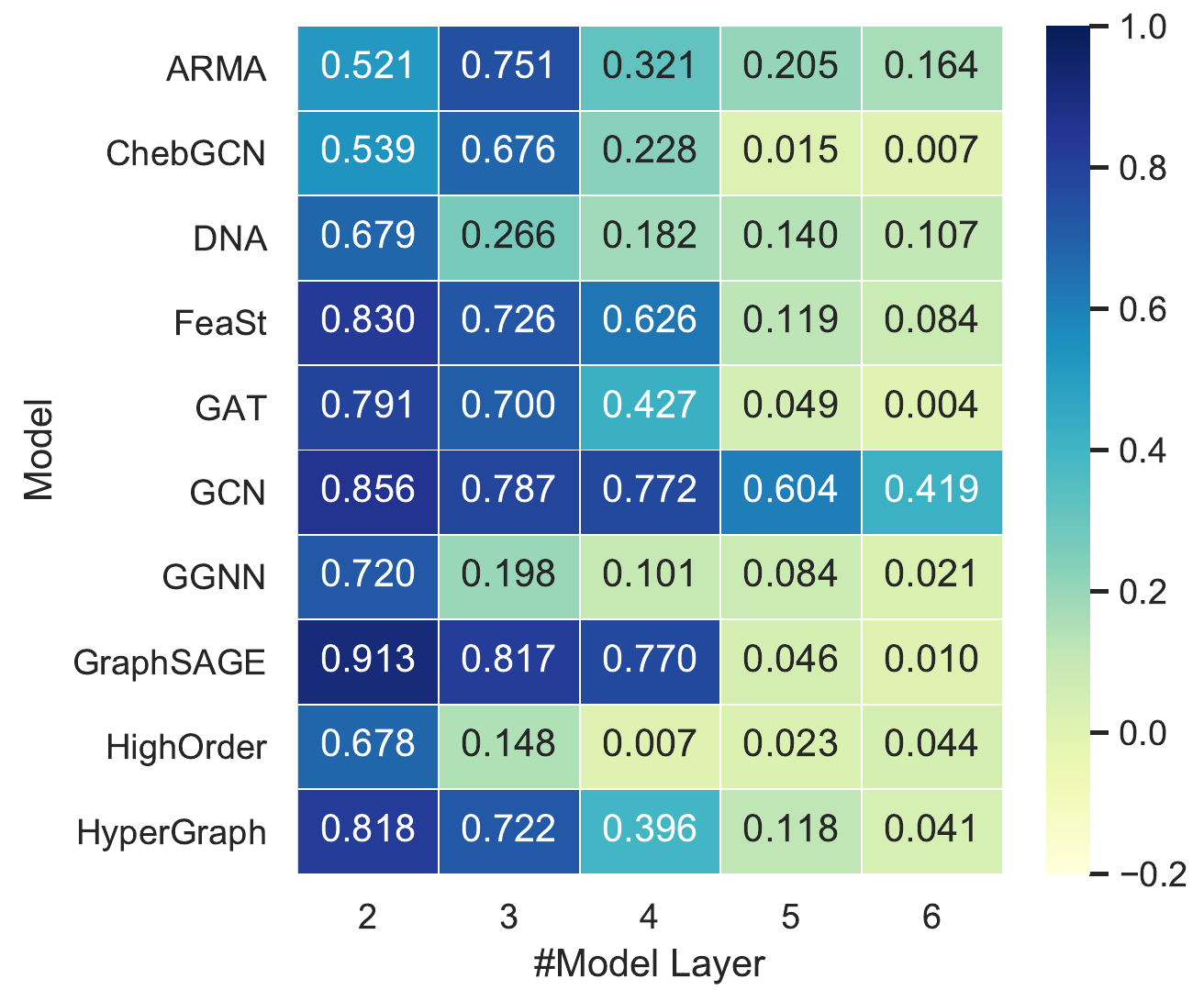}
\end{minipage}%
\\
\begin{minipage}{.48\textwidth}
  \centering
  \includegraphics[width=\linewidth]{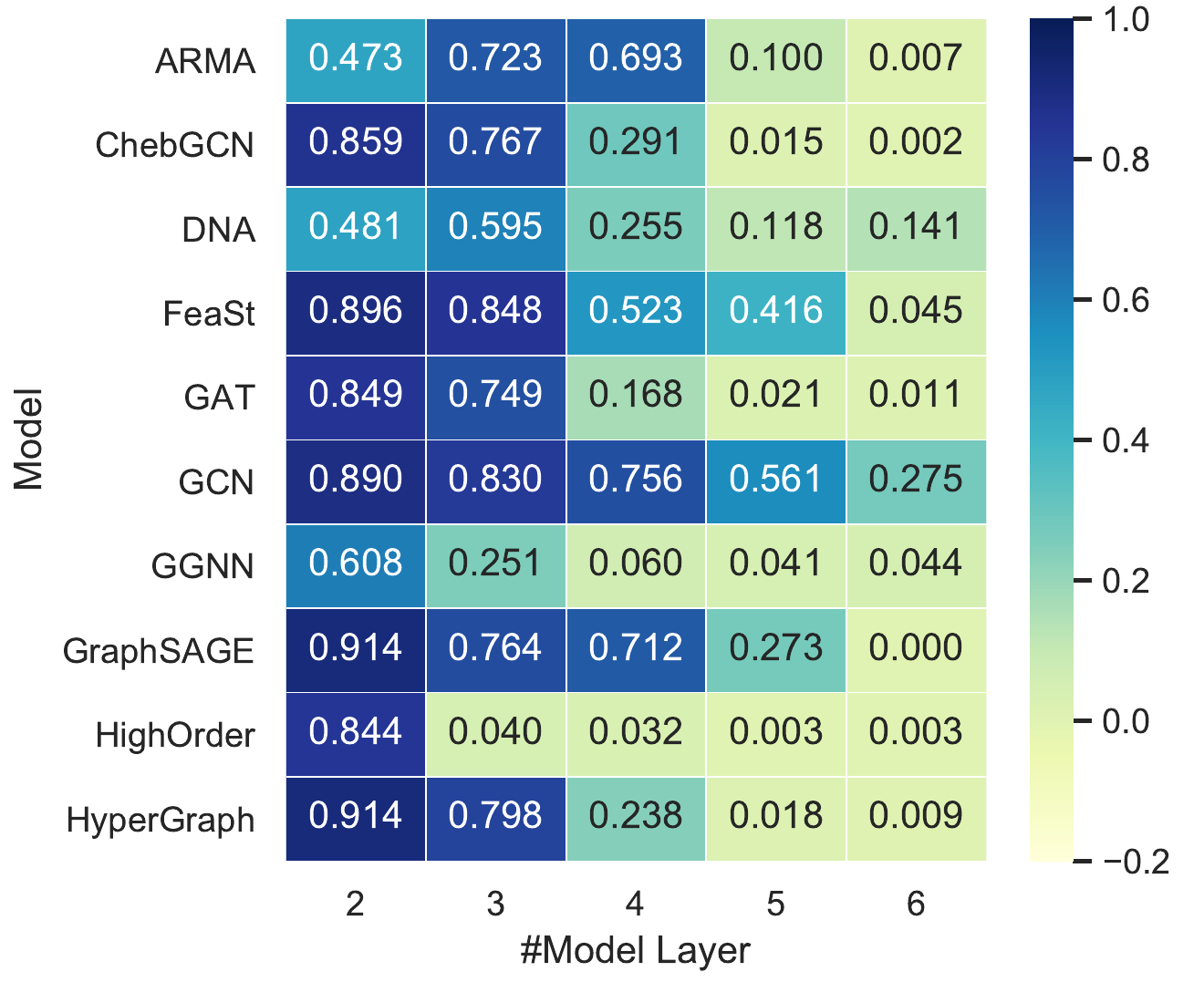}
\end{minipage}%
\caption{MAD results on~\textit{CiteSeer} (top plot) and~\textit{PubMed} (bottom plot) datasets.}
\label{figure_mergeA}
\end{figure}

In Figure~\ref{figure_mergeA}, we display the MAD values of various GNNs with different layers on \textit{CiteSeer} and \textit{PubMed} dataset. We can observe that in these two datasets, the MAD values of all baseline GNN models decreases with the increase of model layer.

\section{C Infomation-to-noise Ratio Experiment on More Datasets }
\begin{figure*}[t]
\centering
\small
\begin{minipage}{1.0\textwidth}
  \centering
  \includegraphics[width=\linewidth]{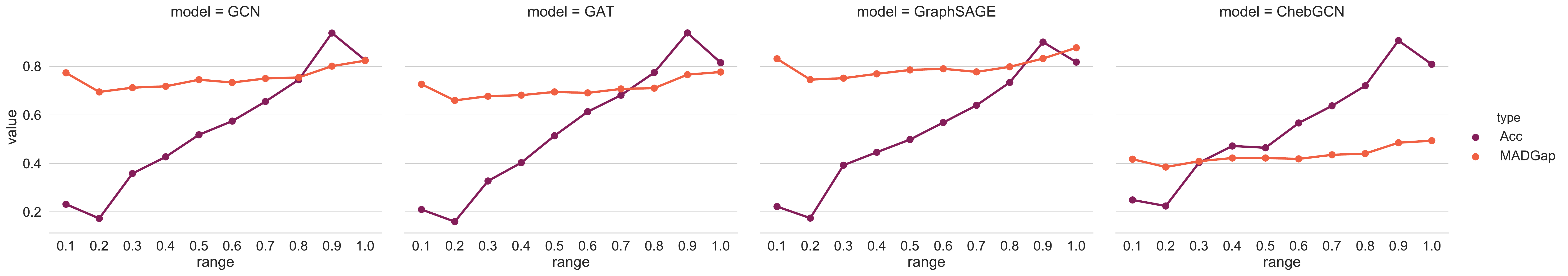}
\end{minipage}%
\\
\begin{minipage}{1.0\textwidth}
  \centering
  \includegraphics[width=\linewidth]{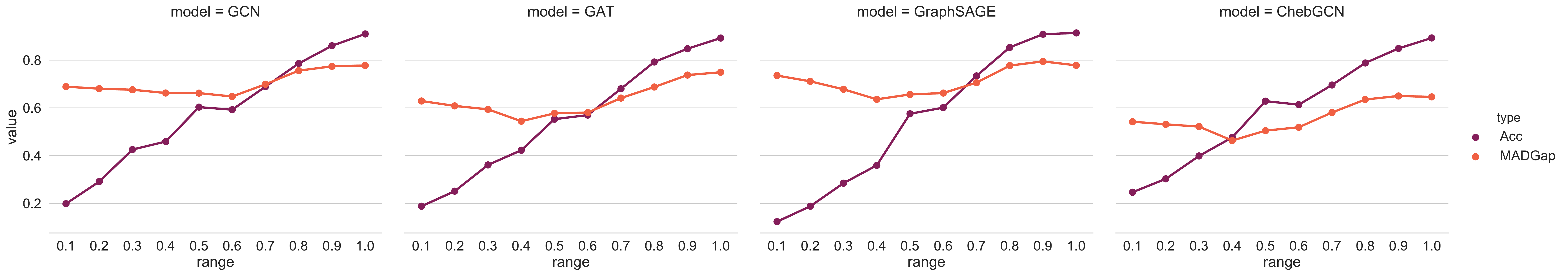}
\end{minipage}%
\caption{Infomation-to-noise Ratio Experiment on~\textit{CiteSeer} (top plot) and~\textit{PubMed} (bottom plot) datasets.}
\label{figure_mergeB}
\end{figure*}

In Figure~\ref{figure_mergeB}, we display the model performance and the MADGap value of node sets with different infomation-to-noise ratio at 2-order neighbours in \textit{CiteSeer} and \textit{PubMed} dataset. We can find that in these four GNN models, the model performance and the MADGap value rises
with the increase of the intra-class node ratio, which will bring more useful information for the graph nodes. So it is the information-to-noise ratio that largely affects the node representation over-smoothness thus has a influence on the model performance.

\section{D AdaEdge Algorithm}
The details of AdaEdge Algorithm is displayed in Algorithm~\ref{loop_training}.
Besides, we use several heuristic approaches to adjust graph topology more wisely: (1) Operation edge for nodes with high prediction confidence (the max value after softmax operation); (2) Operation edge for nodes belong to classes with high prediction precision; (3) Skip operation by a certain probability to control the sparsity of the added and removed edges; (4) Operation edge for nodes with certain degrees.

\section{E Supplementary Result of Relieving Over-smoothing}
\begin{figure*}[!t]
\centering
\includegraphics[scale=0.55]{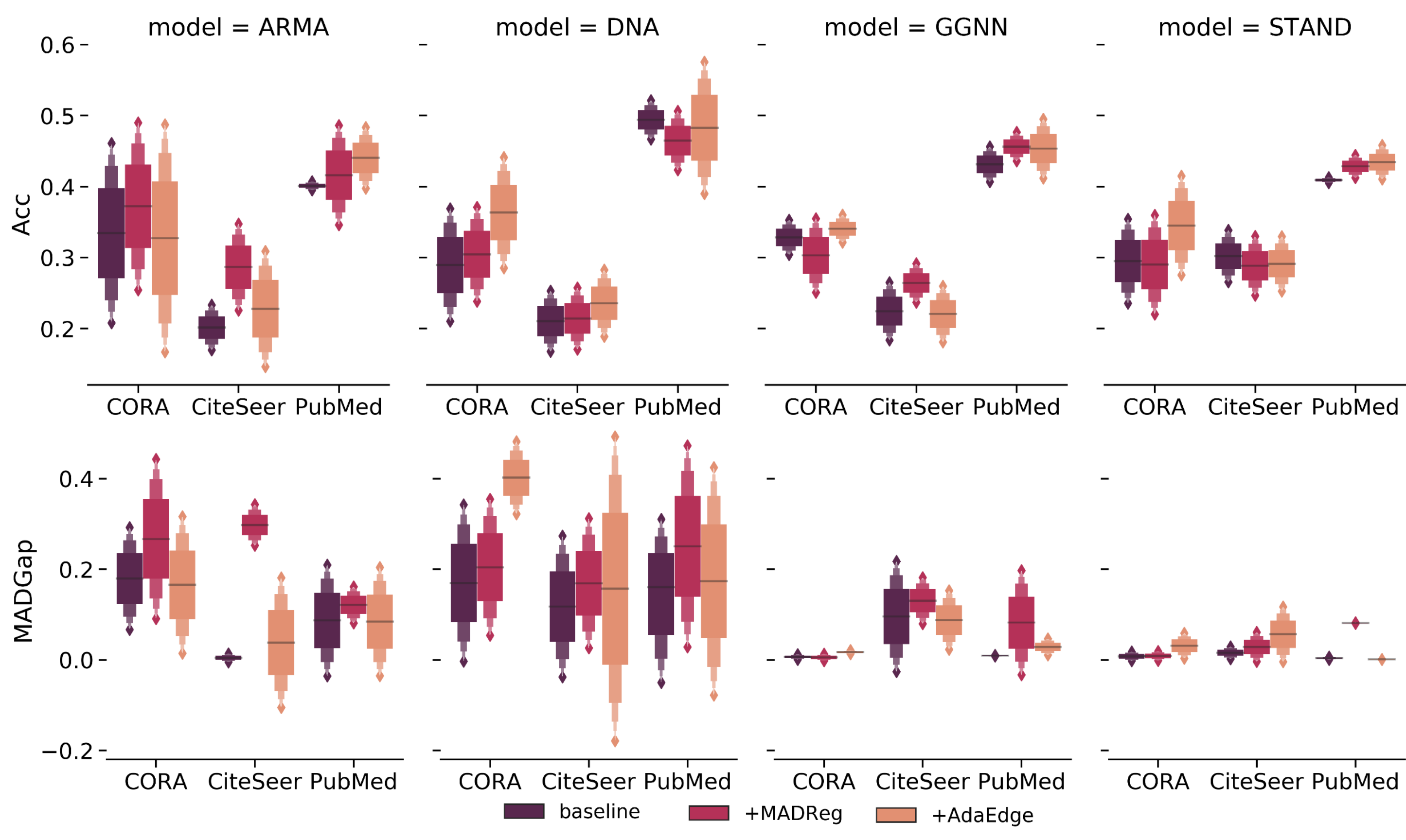}
\caption{More Results on the \textit{CORA/CiteSeer/PubMed} datasets. The number of model layer is 4, where the over-smoothing issue is serious. The box graph shows the mean value and the standard deviation of the prediction accuracy and the MADGap values. And we can find that the two proposed methods can effectively relieve the over-smoothing issue and improve model performance in most cases.}
\label{figure_box2}
\end{figure*}

In Figure~\ref{figure_box2}, we display more models results of relieving over-smoothing in high layer GNNs. 

\section{F Error Analysis of MADReg}
In Table~\ref{table_reg_result2}, we show the results for GCN with different number of layers on the \textit{CORA}/\textit{CiteSeer}/\textit{PubMed} datasets and we observe that the MADReg method can relieve the over-smoothing problem and improve model performance effectively especially for GNNs with more layers where the over-smoothing issue is more severe.
In the shallow GCN, our MADReg method can not effectively improve the model performance because the model has already been well trained and the over-smoothing issue is not serious.

\begin{algorithm}[t]
\centering
\footnotesize
\begin{algorithmic}[1]
\caption{AdaEdge}\label{loop_training}
\Require
The GNN model $\mathbf{GNN_k}$, the feature matrix $\bm{X}$, the raw adjacency matrix $\bm{A}$, the node size $N$,the operation order $order$, the limit number $num_{+}$ and $num_{-}$, the limit confidence $conf_{+}$ and $conf_{-}$ and max training times $max_t$.
\Function{ADDEDGE}{$\mathbf{A},\mathbf{pred},\mathbf{conf}$}
    \State add\_count $\gets$ 0 
    \State $\mathbf{A'} \gets \mathbf{A}$
    \For{node1 $ \mathrm{n_i} \in$  $[0,N)$}
    \For{node2 $ \mathrm{n_j} \in$  $[\mathrm{n_i},N)$}
    \If{$\mathbf{A}_{ij}$==0 and $\mathbf{pred}[\mathrm{n_i}]$==$\mathbf{pred}[\mathrm{n_j}]$ and $\mathbf{conf}[\mathrm{n_i}]$$\ge$$conf_+$ and $\mathbf{conf}[\mathrm{n_j}]$$\ge$$conf_+$  } 
    \State $\mathbf{A'}_{ij} \gets 1$, $\mathbf{A'}_{ji} \gets 1$
    \State add\_count $\gets$ add\_count + 1 
    \If{add\_count $\ge$ $num_+$} \State $\mathbf{reutrn \, A'}$ \EndIf
    \EndIf
    \EndFor
    \EndFor
    \State $\mathbf{reutrn\, A'}$
\EndFunction \\
\Function{REMOVEDGE}{$\mathbf{A},\mathbf{pred},\mathbf{conf}$}
    \State rmv\_count $\gets$ 0 
    \State $\mathbf{A'} \gets \mathbf{A}$
    \For{edge $ (\mathrm{n_i},\mathrm{n_j}) \in$  $\mathbf{A}$}
    \If{$\mathbf{A}_{ij}$==1 and $\mathbf{pred}[\mathrm{n_i}]$!=$\mathbf{pred}[\mathrm{n_j}]$ and $\mathbf{conf}[\mathrm{n_i}]$$\ge$$conf_-$ and $\mathbf{conf}[\mathrm{n_j}]$$\ge$$conf_-$  } 
    \State $\mathbf{A'}_{ij} \gets 0$, $\mathbf{A'}_{ji} \gets 0$
    \State rmv\_count $\gets$ rmv\_count + 1 
    \If{rmv\_count $\ge$ $num_-$} \State $\mathbf{reutrn \, A'}$ \EndIf
    \EndIf
    \EndFor
    \State $\mathbf{reutrn\, A'}$
\EndFunction \\
\Function{ADJUSTGRAPH}{$\mathbf{pred},\mathbf{conf}$}
\If{$order$=='add\_first'}
\State $\mathbf{A'} \gets \mathrm{ADDEDGE} (\bm{A},\mathbf{pred},\mathbf{conf})$
\State $\mathbf{A'} \gets \mathrm{REMOVEEDGE} (\mathbf{A'},\mathbf{pred},\mathbf{conf})$
\Else
\State $\mathbf{A'} \gets \mathrm{REMOVEEDGE} (\bm{A},\mathbf{pred},\mathbf{conf})$
\State $\mathbf{A'} \gets \mathrm{ADDEDGE} (\mathbf{A'},\mathbf{pred},\mathbf{conf})$
\EndIf
\State $\mathbf{reutrn\, A'}$
\EndFunction \\

\State \textbf{AdaEdge}
\State $\mathrm{acc_0},\mathbf{pred_0},\mathbf{conf_0} \gets \mathbf{GNN_k^0}(\bm{A},\bm{X})$
\State $\mathbf{A'_0} \gets \mathrm{ADJUSTGRAPH}(\mathbf{pred_0},\mathbf{conf_0})$
\For{iter times $ \mathrm{i} \in [1,max_t)$ }
\State $\mathrm{acc_i},\mathbf{pred_i},\mathbf{conf_i} \gets \mathbf{GNN_k^i}(\mathbf{A'_{i-1}},\bm{X})$
\If{$\mathrm{acc_i} \le \mathrm{acc_{i-1}}$}
\State $\mathbf{reutrn\,} \mathrm{acc_{i-1}} $
\EndIf
\State $\mathbf{A'_i} \gets
\mathrm{ADJUSTGRAPH}(\mathbf{pred_i},\mathbf{conf_i})$
\EndFor
\State $\mathbf{reutrn\,} \mathrm{acc_{max_t-1}} $
\end{algorithmic}
\end{algorithm}

\end{document}